\documentclass[conference]{IEEEtran}
\IEEEoverridecommandlockouts
\usepackage{cite}
\usepackage{amsmath,amssymb,amsfonts}
\usepackage{algorithmic}
\usepackage{graphicx}
\usepackage{textcomp}
\usepackage{xcolor}
\usepackage{url}
\def\BibTeX{{\rm B\kern-.05em{\sc i\kern-.025em b}\kern-.08em
    T\kern-.1667em\lower.7ex\hbox{E}\kern-.125emX}}
\begin{document}

\title{Albumentations: fast and flexible image augmentations}


\author{\IEEEauthorblockN{Alexander Buslaev}
\IEEEauthorblockA{\textit{MapBox} \\
Minsk, Belarus \\
al.buslaev@gmail.com}
\and
\IEEEauthorblockN{Alex Parinov}
\IEEEauthorblockA{\textit{Rick.ai} \\
Samara, Russia \\
creafz@gmail.com}
\and
\IEEEauthorblockN{Eugene Khvedchenya}
\IEEEauthorblockA{\textit{ODS.ai} \\
Odessa, Ukraine \\
ekhvedchenya@gmail.com}
\and
\IEEEauthorblockN{Vladimir I. Iglovikov}
\IEEEauthorblockA{\textit{Level5 Engineering, Lyft Inc} \\
Palo Alto, CA, USA \\
iglovikov@gmail.com}
\and
\IEEEauthorblockN{Alexandr A. Kalinin}
\IEEEauthorblockA{\textit{University of Michigan} \\
Ann Arbor, MI, USA \\
akalinin@umich.edu}
}

\maketitle

\begin{abstract}
Data augmentation is a commonly used technique for increasing both the size and the diversity of labeled training sets by leveraging input transformations that preserve output labels.
In computer vision domain, image augmentations have become a common implicit regularization technique to combat overfitting in deep convolutional neural networks and are ubiquitously used to improve performance.
While most deep learning frameworks implement basic image transformations, the list is typically limited to some variations and combinations of flipping, rotating, scaling, and cropping.
Moreover, the image processing speed varies in existing tools for image augmentation.
We present Albumentations, a fast and flexible library for image augmentations with many various image transform operations available, that is also an easy-to-use wrapper around other augmentation libraries. We provide examples of image augmentations for different computer vision tasks and show that Albumentations is faster than other commonly used image augmentation tools on the most of commonly used image transformations.
The source code for Albumentations is made publicly available online at \url{https://github.com/albu/albumentations}.
\end{abstract}


\section{Introduction}
Modern machine learning models, such as deep artificial neural networks, often have a very large number of parameters, which allows them to generalize well when trained on massive amounts of labeled data.
In practice, such large labeled datasets are not always available for training, which leads to the elevated risk of overfitting.
Data augmentation is a commonly used technique for increasing both the size and the diversity of labeled training sets by leveraging input transformations that preserve output labels. In computer vision domain, image augmentations have become a common implicit regularization technique to combat overfitting in deep convolutional neural networks and are ubiquitously used to improve performance on benchmark datasets \cite{Simard2003bestpractices,krizhevsky2012imagenet,howard2013some,ratner2017learning,cubuk2018autoaugment}.

While most deep learning frameworks implement basic image transformations, the list is typically limited to some variations and combinations of flipping, rotating, scaling, and cropping.
Different domains, imaging modalities, and tasks may benefit from a wide range of different amounts and combinations of image transformations \cite{ratner2017learning,lemley2017smartaug,cubuk2018autoaugment}.
For example, very extensive image augmentations are typical in medical image analysis, where datasets are often small and expensive to acquire, and annotations are often sparse, compared to the natural images \cite{ching2017opportunities,rakhlin2018deep}.
Even on well studied benchmark datasets, different data augmentation strategies lead to performance variability.
For example, image rotation is an effective data augmentation method on CIFAR-10, but not on MNIST, where it can negatively affect the network's ability to distinguish between handwritten digits \textit{6} and \textit{9} \cite{lemley2017smartaug}. 
Thus, there is a need for flexible and rich image augmentation tools that will allow to apply combinations of a wide range of various transformations to the task in hand.

While the largest gains in efficiency of deep neural networks have come from computation with graphical processing units (GPUs) \cite{edwards2015growing}, which excel at the matrix and vector operations central to deep learning, data pre-processing and augmentation are typically done on a CPU.
As the performance and the memory capacity of the GPU hardware steadily improve, the efficiency of data augmentation operations becomes increasingly important, such that GPUs are not idling while CPUs are preparing the next mini-batch of data to pass through the network.
However, the image processing speed varies in existing tools for image augmentation.

In this paper we present Albumentations, a fast and flexible solution for image augmentations. Albumentations is a library based on a fast implementations of a large number of various image transform operations, but also is an easy-to-use wrapper around other augmentation libraries. It provides a simple yet powerful interface for different tasks, including image classification, segmentation, detection, etc. We demonstrate that Albumentations is faster than other commonly used image augmentation tools on the most of commonly used image transformations.


\section{Albumentations}
\begin{figure}[h]
\includegraphics[width=\linewidth]{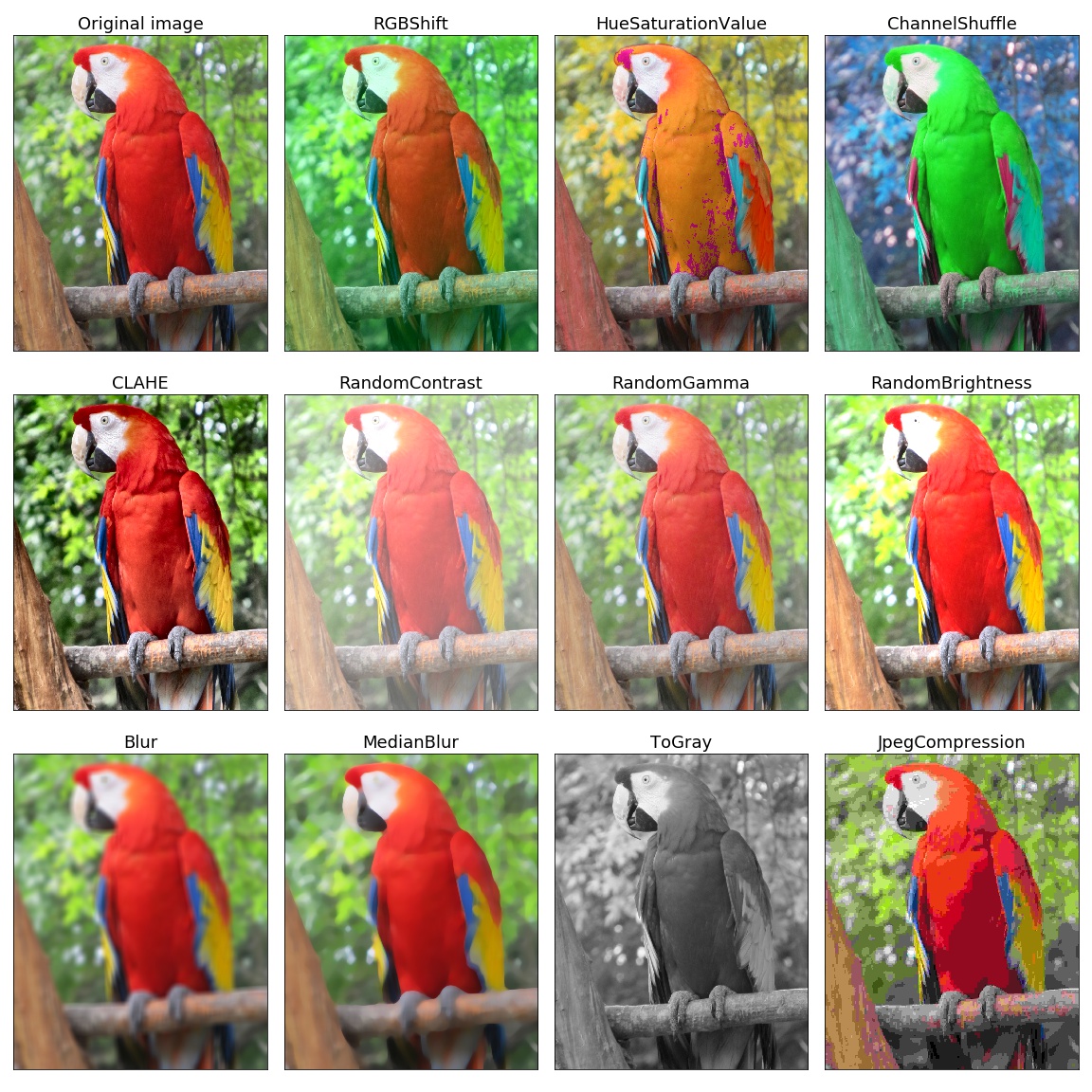}
\caption{Exemplar applications of image transformations available in Albumentations}
\label{fig::parrot}
\end{figure}

Fig. \ref{fig::parrot} demonstrates application of a number of image transformation operations available in Albumentations, including color, contrast, brightness, and other transformations. These image operations and their various combinations are commonly used in a wide range of computer vision tasks. For this example, we selected the parameters of transformation such that results of their applications visibly differ from the original image, which may not be always desirable in practice.

Below we provide task-specific augmentation examples for different computer vision tasks.

\subsection{Street view image detection}
\begin{figure}[h]
\includegraphics[width=\linewidth]{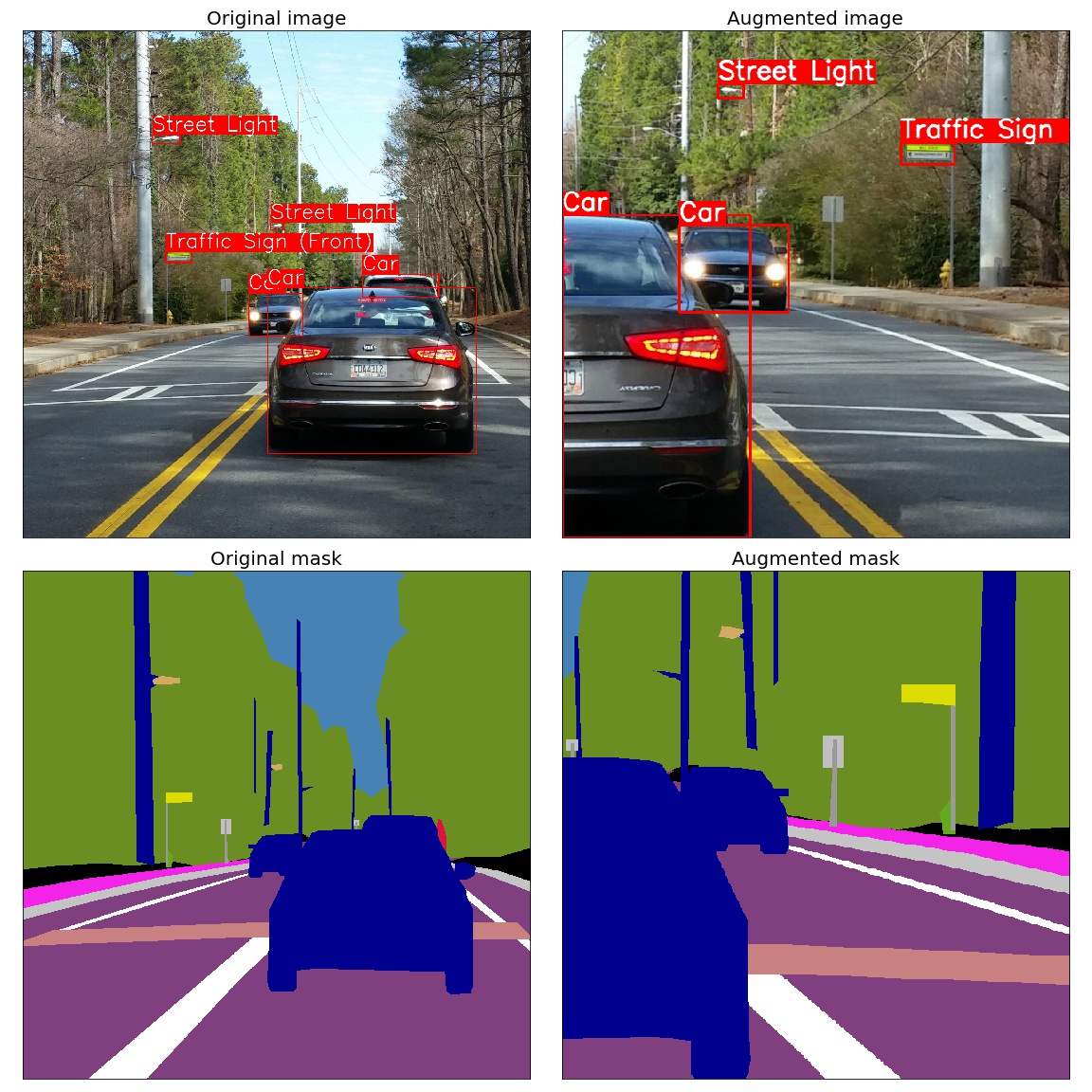}
\caption{An example of applying a combination of transformations available in Albumentations to the original image, bounding boxes, and ground truth masks for instance segmentation.}
\label{fig::vistas}
\end{figure}

Computer vision tasks are not limited to image classification, thus, the support of augmentations for other data formats is also important. Many spacial transforms implemented in Albumentations support operating on segmentation masks and bounding boxes, that are used in object detection and tracking.сп
Fig. \ref{fig::vistas} shows an example of applying a combination of a horizontal flip and a random sized crop to an image from the Mapillary Vistas Dataset for Semantic Understanding of Street Scenes \cite{neuhold2017mapillary}. Besides the original image, it shows the result of transform application on both bounding box and instance mask annotations.

\subsection{Satellite and aerial imagery}
\begin{figure*}[h]
\includegraphics[width=\linewidth]{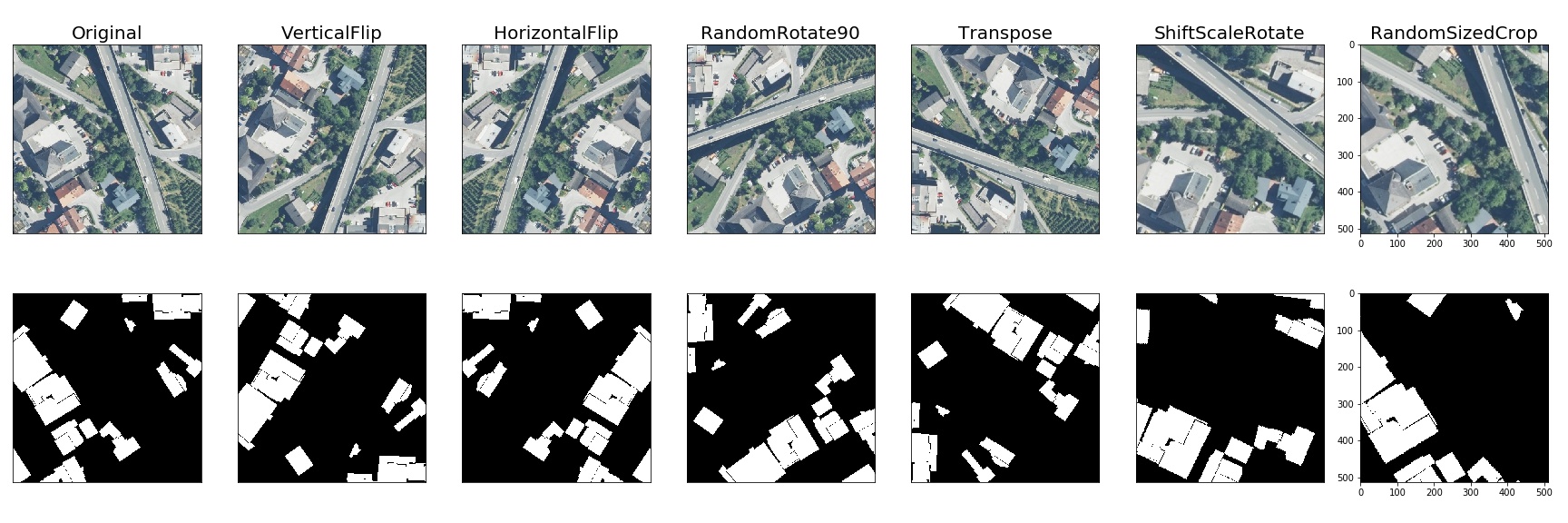}
\caption{Exemplar geometry-preserving transforms applied to a satellite image (top row) and ground truth binary segmentation mask (bottom row).}
\label{fig::inria}
\end{figure*}

In the analysis of satellite and aerial images, transformations that preserve the shape of objects are typically used to avoid the distortions of rigid-shape objects, such as buildings.
Such transform operations include cropping, rotations, reflections, and scaling.
For example, on of the top-3 performing solutions in the DSTL Satellite Imagery Feature Detection challenge on Kaggle used combinations of random cropping and a random transformation from dihedral group $Dih_4$ \cite{iglovikov2017satellite}.
High-performing solutions to other challenges also implemented similar types of image augmentations, for example, for automatic road extraction \cite{buslaev2018fully} and for multi-class land segmentation \cite{seferbekov2018feature} from satellite imagery.
Fig. \ref{fig::inria} provides examples of such image transforms applied to an image from the Inria Aerial Image Labeling dataset \cite{maggiori2017can}. Corresponding binary masks of buildings present in the image are also transformed in the same way as the original image.

\subsection{Biomedical image analysis}
Image augmentations are intensively used in the analysis of biological and medical images due to the typically limited amount of available labeled data \cite{ching2017opportunities}.
For example, combining pre-trained deep network architectures with multiple augmentation techniques enabled accurate detection of breast cancer from a very small set of histology images with less than 100 images per class \cite{rakhlin2018deep}.
Similarly, the use of medical image augmentations helped to improve the results of segmentation of hand radiographs and bone age assessment \cite{iglovikov2017pediatric}.
In medical computer vision tasks that deal with color images or videos, color transformations have also been shown to help deep networks to generalize better, for example, such for surgical \cite{shvets2018automatic} or endoscopic \cite{shvets2018angiodysplasia} video analysis.
In addition to these commonly used transforms, operations like grid distortion and elastic transform often can be helpful, see Fig. \ref{fig::inria}, since medical imaging is often dealing with non-rigid structures that have shape variations.

\begin{figure}[h]
\includegraphics[width=\linewidth]{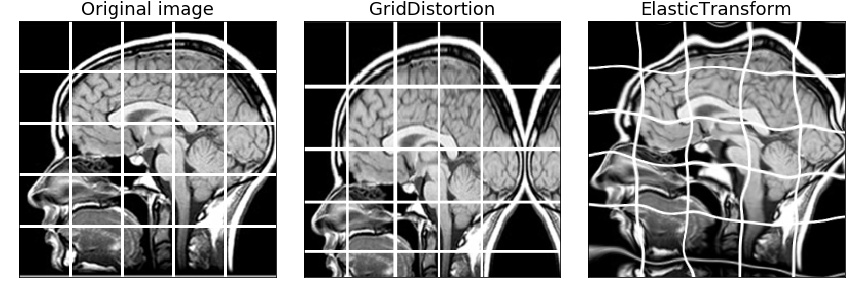}
\caption{Grid distortion and elastic transform applied to a medical image.}
\label{fig::medical}
\end{figure}



\section{Benchmarks}
The quantitative comparison of image transformation speed performance for Albumentations and other commonly used image augmentation tools is presented in Table \ref{table:speed}. We included a framework-agnostic image augmentation library imgaug \cite{jung2017imgaug}, as well as augmentations provided within Keras \cite{chollet2015keras} and PyTorch \cite{paszke2017automatic} frameworks. For the most image operations, Albumentations is consistently faster than all alternatives.

\begin{table*}[th!]
\caption{Time in seconds per image transformation operation task using different augmentation tools (lower is better).}
\label{table:speed}
\centering   
\begin{tabular}{|c | c | c | c | c | c|}
\hline\noalign{\smallskip}
& Albumentations & imgaug & torchvision (Pillow backend) & torchvision (Pillow-SIMD backend) & Keras \\
\hline
RandomCrop64 & \textbf{0.0017} & - & 0.0182 & 0.0182 & -\\
PadToSize512 & \textbf{0.2413} & - & 2.493 & 2.3682 & -\\
HorizontalFlip & 0.7765 & 2.2299 & \textbf{0.3031} & 0.3054 & 2.0508\\
VerticalFlip & \textbf{0.178} & 0.3899 & 0.2326 & 0.2308 & 0.1799\\
Rotate & 3.8538 & 4.0581 & 16.16 & 9.5011 & 50.8632\\
ShiftScaleRotate & \textbf{2.0605} & 2.4478 & 18.5401 & 10.6062 & 47.0568\\
Brightness & \textbf{2.1018} & 2.3607 & 4.6854 & 3.4814 & 9.9237\\
ShiftHSV & \textbf{10.3925} & 14.2255 & 34.7778 & 27.0215 & -\\
ShiftRGB & 2.6159 & \textbf{2.1989} & - & - & 3.0598\\
Gamma & 1.4832 & - & \textbf{1.1397} & 1.1447 & -\\
Grayscale & \textbf{1.2048} & 5.3895 & 1.6826 & 1.2721 & -\\
\hline
\end{tabular}
\end{table*}

\section{Conclusions}
We present Albumentations, a fast and flexible library for image augmentations with many various image transform operations available, that is also an easy-to-use wrapper around other augmentation libraries. We provide examples of image augmentations for different computer vision tasks ans show that Albumentations is faster than other commonly used image augmentation tools on the most of commonly used image transformations.
The source code for Albumentations is made publicly available online at \url{https://github.com/albu/albumentations}.

\bibliographystyle{IEEEtran}
\bibliography{paper}

\end{document}